\documentclass[10pt,twocolumn,letterpaper]{article}

\usepackage{wacv}
\usepackage{times}
\usepackage{epsfig}
\usepackage{graphicx}
\usepackage{amsmath}
\usepackage{amssymb}
\usepackage{multirow}
\usepackage{multicol}
\usepackage{balance}
\usepackage[ruled,linesnumbered]{algorithm2e}
\usepackage{array}
\newcolumntype{H}{>{\setbox0=\hbox\bgroup}c<{\egroup}@{}}
\usepackage{sidecap, caption}
\usepackage{overpic}
\usepackage{algpseudocode}

\usepackage{longtable}
\usepackage{supertabular}
\usepackage{pifont}
\newcommand{\cmark}{\ding{51}}%
\newcommand{\xmark}{\ding{55}}%

\usepackage[dvipsnames]{xcolor}
\definecolor{Orange}{RGB}{255,127,80}
\definecolor{Green}{RGB}{0,128,0}
\definecolor{Red}{RGB}{220,20,60}
\definecolor{Blu}{RGB}{0,0,0}

\def\wacvPaperID{1192} 

\wacvfinalcopy 

\ifwacvfinal
\def\assignedStartPage{1} 
\fi

\ifwacvfinal
\usepackage[breaklinks=true,bookmarks=false]{hyperref}
\else
\usepackage[pagebackref=true,breaklinks=true,colorlinks,bookmarks=false]{hyperref}
\fi

\ifwacvfinal
\else
\fi

\begin{document}

\title{Learning without Seeing nor Knowing: Towards Open Zero-Shot Learning}

\author{Federico Marmoreo$^{1,2}$, Julio Ivan Davila Carrazco$^{1,2,5}$, Vittorio Murino$^{1,3,4}$, Jacopo Cavazza$^{1,5}$\\
$^1$Pattern Analysis and Computer Vision (PAVIS), Istituto Italiano di Tecnologia, Italy \\
$^2$Universit\`a degli Studi di Genova, Italy  \\
$^3$Huawei Technologies Ltd., Ireland Research Center, Ireland \\
$^4$Department of Computer Science, University of Verona, Italy \\
$^5$Visual Geometry and Modelling (VGM), Istituto Italiano di Tecnologia, Italy \\
{\tt\small \{federico.marmoreo,julio.davila,vittorio.murino,jacopo.cavazza\}\@iit.it}}

\maketitle

\begin{abstract}
In Generalized Zero-Shot Learning (GZSL), unseen categories (for which no visual data are available at training time) can be predicted by leveraging their class embeddings (\eg, a list of attributes describing them) together with a complementary pool of seen classes (paired with both visual data and class embeddings). Despite GZSL is arguably challenging, we posit that knowing in advance the class embeddings, especially for unseen categories, is an actual limit of the applicability of GZSL towards real-world scenarios. To relax this assumption, we propose Open Zero-Shot Learning (OZSL) to extend GZSL towards the open-world settings. We formalize OZSL as the problem of recognizing seen and unseen classes (as in GZSL) while also rejecting instances from unknown categories, for which neither visual data nor class embeddings are provided. We formalize the OZSL problem introducing evaluation protocols, error metrics and benchmark datasets. We also suggest to tackle the OZSL problem by proposing the idea of performing unknown feature generation (instead of only unseen features generation as done in GZSL). We achieve this by optimizing a generative process to sample unknown class embeddings as complementary to the seen and the unseen. We intend these results to be the ground to foster future research, extending the standard closed-world zero-shot learning (GZSL) with the novel open-world counterpart (OZSL). 
\end{abstract}

\section{Introduction}

\begin{figure}[t!]
    \centering
    \includegraphics[width=\columnwidth]{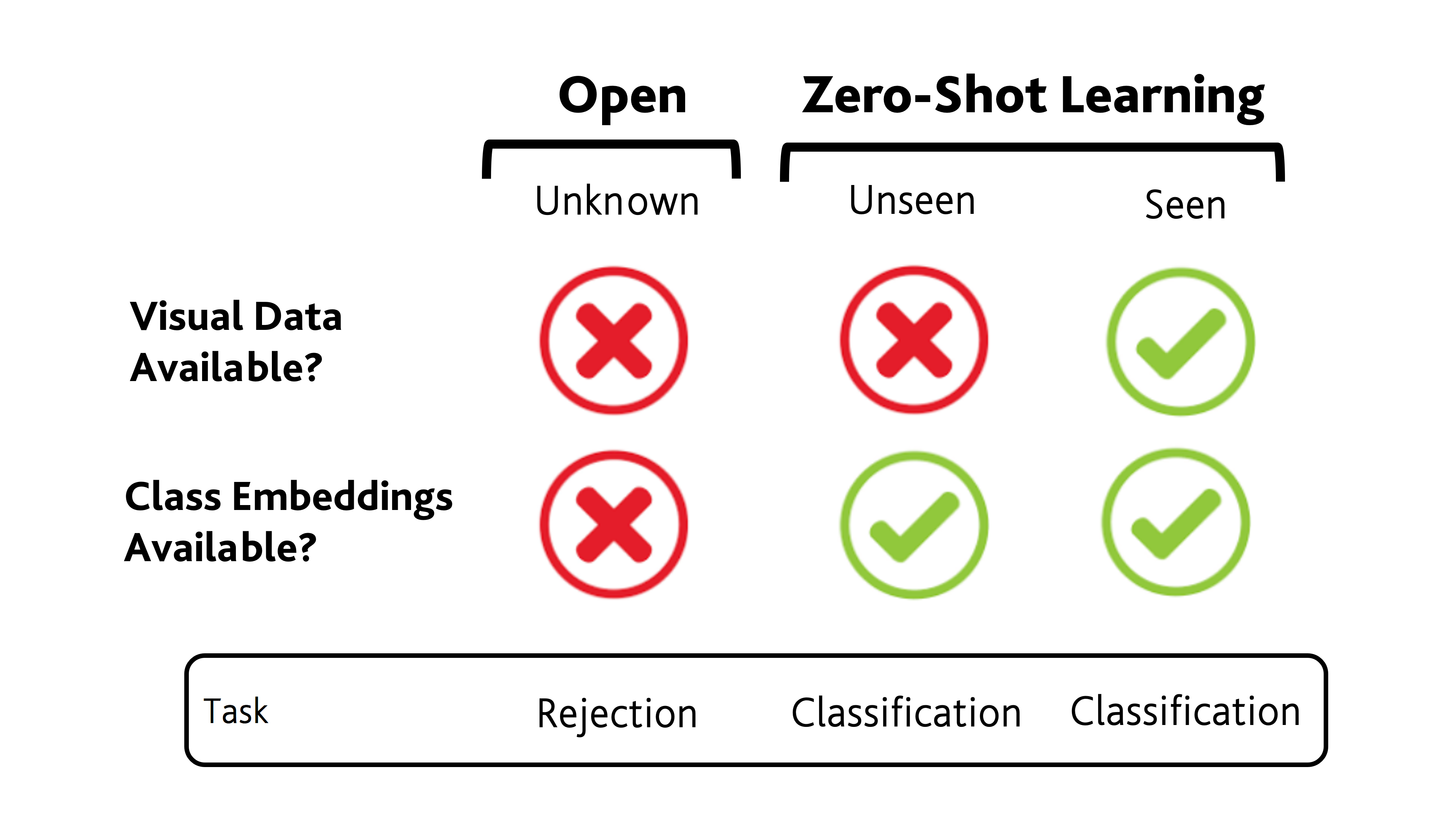}
    \caption{{\bf Open Zero-Shot Learning}, a framework where we aim at classifying seen and unseen classes (for which no visual data of the latter is given) while also rejecting (\ie, refusing to take any decision on) unknown classes. Neither visual data nor class embeddings are available for unknown classes.
} 

    \label{fig:OZSL}
\end{figure}

After the advent of deep learning and related end-to-end-trainable architectures, computer vision has reached near human-level performance on a variety of tasks. However, the main operative assumption behind this outstanding performance is the availability of a large corpus of annotated data and this clearly limits the applicability in a real-world scenario. Generalized Zero-Shot Learning (GZSL) \cite{ChaoEmpGZSL} considers the extreme case in which for some of the classes, \ie, the \textit{unseen} classes, no training examples are available. The goal is to correctly classify them at inference time, together with test instances from the seen classes, and this is typically achieved relying on auxiliary semantic (\eg, textual) information describing the classes, the so-called \textit{class embeddings} \cite{lampert2009learning}.

\begin{SCfigure*}[][t!]
    \centering
    \includegraphics[width=0.75\textwidth]{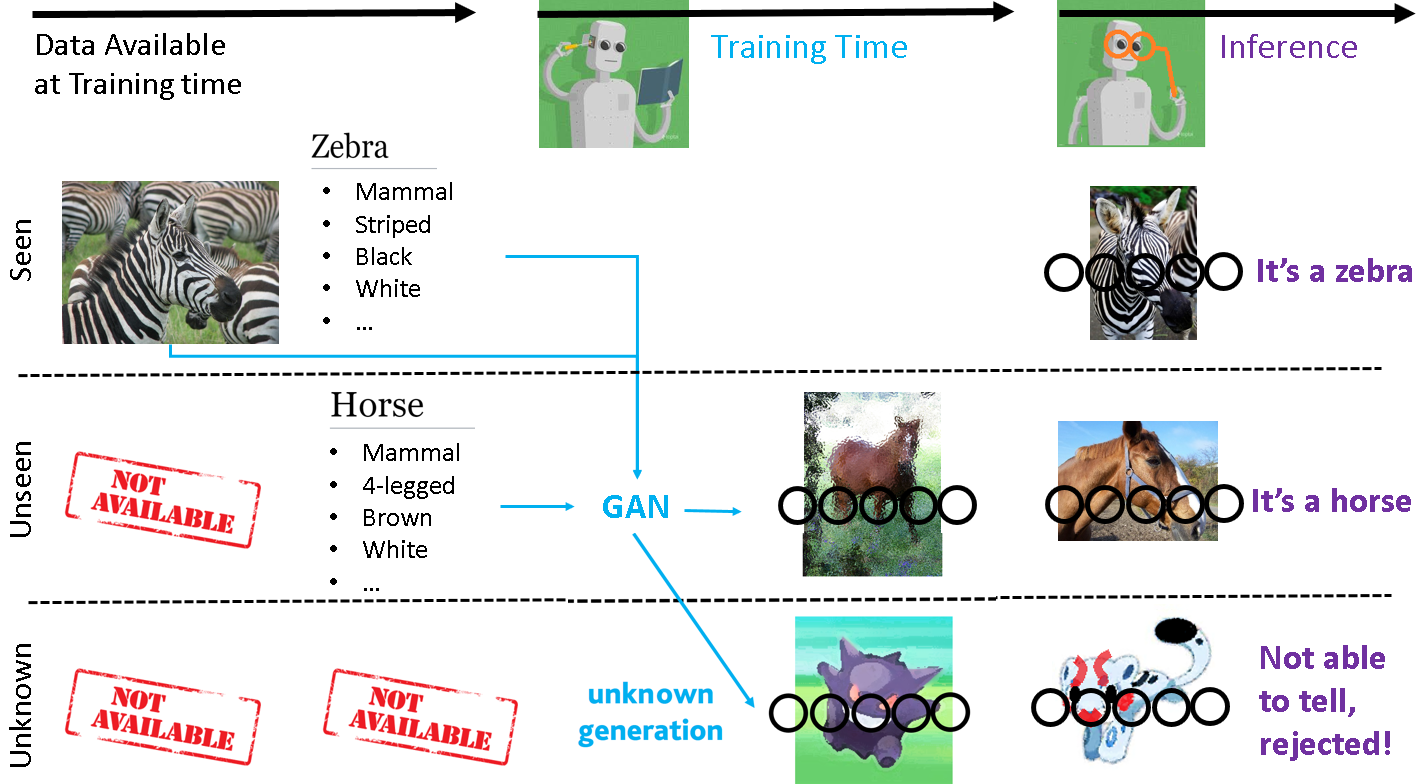}
    \caption{{\bf The proposed pipeline for Open Zero-Shot Learning (OZSL).} We synthesize visual descriptors from seen and unseen classes, using a Generative Adversarial Network (GAN). We also learn how to perform unknown generation and synthesize descriptors (represented by \protect\includegraphics[width=1cm]{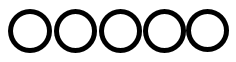}), even for the unknown classes, and better precondition a classifier in classifying seen/unseen and reject unknown, with the usage of Openmax \cite{OpenMax}.}
    \label{fig:my_label}
\end{SCfigure*}

For instance, class embeddings can either consist of side information such as manually-defined attributes 
codified by Osherson's default probability scores \cite{lampert2009learning}, text embeddings extracted from computational pipelines such as word2vec \cite{mikolov2013distributed}, or CNN+LSTM models trained on Wikipedia articles \cite{Xian_2019_CVPR}. 
Desirable features of class embeddings consist in being 1) shared among classes and, at the same time, 2) discriminative. This is how one can transfer knowledge from the classes for which we have annotated visual data, \ie the \textit{seen} classes, to the unseen ones.

In GZSL, the challenge is to overcome the bias of the model towards predicting the classes on which it has been directly trained on, and for which it is much more confident in forecasting. To solve the extreme imbalance of the GZSL framework, much effort has been exerted to perform synthetic feature augmentation for the unseen classes  \cite{mishra2018generative,Arora2018GeneralizedZL,felix2018multi,huang2019generative,Li:CVPR19,Xian_2019_CVPR,Schonfeld_2019_CVPR,Zhu:ICCV19,Xian_2018_CVPR,Sariyildiz2019GradientMG,Gao2020ZeroVAEGANGU,Han_2021_CVPR}.
By exploiting deep generative models, as Generative Adversarial Networks (GANs) or Variational Auto-Encoders (VAEs), it is indeed possible to take advantage of the class embeddings to generate class consistent features for the unseen classes by training on the seen ones, leading to remarkable performances in GZSL.

However, we claim that the assumption of knowing in advance the full set of classes, the closed-world assumption, and their class embeddings is still a strong limitation for GZSL in real world applications. In fact, while it is reasonable to assume that we can describe all the seen classes with the class embeddings, it seems less reasonable not only to know, but also to describe with the rich semantic content of the class embeddings, all the classes for which we have no visual training data.
  
\paragraph{\textit{We introduce a new paradigm, OZSL}} (Figure \ref{fig:OZSL}). \textit{Open Zero-Shot learning} overcomes the closed-world assumption and goes to the open-world scenario by considering a possible infinite set of classes at inference time. As a consequence, we have three types of classes: 1) \textit{the seen}, for which we have visual data and class semantic descriptors, 2) \textit{the unseen}, for which we have only class embeddings, and 3) \textit{the unknown}, for which we have neither the visual data nor the (semantic) class embeddings. Thus, OZSL extends GZSL with the possibility of performing recognition in the open-set regime \cite{Scheirer2012} where inference 
has to be jointly performed over seen, unseen and unknown classes in order to classify seen and unseen, and reject unknown ones. 

\paragraph{\textit{We build OZSL as the open-world generalization of GZSL.}} To warm up the research community towards the solution of OZSL, we design evaluation protocols, extracting unknown classes as a subpart of unseen classes from typical GZSL benchmark datasets used in the related state of the art \cite{mishra2018generative,Arora2018GeneralizedZL,felix2018multi,huang2019generative,Li:CVPR19,Xian_2019_CVPR,Schonfeld_2019_CVPR,Zhu:ICCV19,Xian_2018_CVPR,Sariyildiz2019GradientMG,Gao2020ZeroVAEGANGU,Han_2021_CVPR}. We will make these splits publicly available so as to ease the research community in this direction, and we also propose error metrics to allow fair and reproducible comparison across different algorithmic solutions tackling OZSL. 
We also extend prior GZSL error metrics (harmonic mean of the per-class average accuracy \cite{xian2018tPAMI}) to better handle the open world scenario. In particular, we consider F1-score between seen and unseen average precision and/or recall scores to better account for successful rejections.

\paragraph{\textit{We approach OZSL by synthesizing the unknown.}} (Figure \ref{fig:my_label}). In GZSL, GANs or alternative generative methods \cite{mishra2018generative,Arora2018GeneralizedZL,felix2018multi,huang2019generative,Li:CVPR19,Xian_2019_CVPR,Schonfeld_2019_CVPR,Zhu:ICCV19,Xian_2018_CVPR,Sariyildiz2019GradientMG,Gao2020ZeroVAEGANGU,Han_2021_CVPR}) generate visual features conditioned on class embeddings in order to synthesize descriptors for the unseen classes and train a softmax classifier on top of them as well as of real seen features. 
We purport that we can easily extend this state-of-the-art paradigm to OZSL by replacing the standard softmax classifier with Openmax\footnote{Openmax \cite{OpenMax} augments the softmax classes' bins (out of which probabilities are arg-maxed to compute predictions) by introducing an extra-bin estimating the probability to reject an instance. Thanks to Openmax, we can still cast recognition over seen \& unseen classes, and rejection over unknown classes through a single $\arg\max$ step.} \cite{OpenMax}. We provide a preliminary exploratory analysis, evaluating both baseline methods (\eg, GZSL feature generator simply borrowed for OZSL) and our novel idea to synthesize unknown class embeddings and using them to generate unknown visual features, 
which we implemented through a variation of Wasserstein GANs \cite{improvedWGAN,Xian_2018_CVPR,narayan20}, which we term VAcWGAN (variationally-conditioned Wasserstein GAN). VAcWGAN optimizes a conditional generative process on semantic embeddings (so that, we first ``synthesize the unknown'' and then we generate unknown visual features). Despite this approach being arguably harder (since we attempt to generate something we do not see nor know), our experimental evidence shows some potential which we deem worthy to be further investigated by the computer vision community.

\paragraph{Outline of the paper.} Sect. \ref{sez:rl} addresses the related works and highlights the new introduced problem and the main novel aspects of the method proposed to tackle it. Sect. \ref{sez:OZSL} formalizes the new OZSL problem and introduces benchmarks and performance metrics. In Sect. \ref{sez:method}, the proposed method (VAcWGAN) is reported. Sect. \ref{sez:exp} presents our experimental analysis (proposing baseline models, ablation studies and instantiating the new SOTA for OZSL). Finally, Sect. \ref{sez:conc} draws conclusions and sketches possible future work.

\section{Related work}\label{sez:rl}

\paragraph{Generalized Zero-Shot Learning.} Feature generating networks are surely a ``big thing'' for GZSL \cite{mishra2018generative,Arora2018GeneralizedZL,felix2018multi,huang2019generative,Li:CVPR19,Xian_2019_CVPR,Schonfeld_2019_CVPR,Zhu:ICCV19,Xian_2018_CVPR,Sariyildiz2019GradientMG,Gao2020ZeroVAEGANGU,Han_2021_CVPR}. As proposed by \cite{Xian_2019_CVPR} and \cite{zhu2018generative} almost independently, a (Wasserstein) GAN, conditioned on class embeddings, is paired with a classification loss in order to generate sufficiently discriminative CNN features, which are then fed to a softmax classifier for the final inference stage.

Recently, several modifications have been adopted to improve feature generation for ZSL,
for instance, by either replacing the GAN with a variational autoencoder \cite{mishra2018generative,Arora2018GeneralizedZL} or  using the latter two models in parallel \cite{Xian_2019_CVPR,Gao2020ZeroVAEGANGU}, cycle consistency loss \cite{felix2018multi,huang2019generative} or contrastive loss \cite{Han_2021_CVPR}. In \cite{Li:CVPR19}, class embeddings are regressed from visual features, while semantic-to-visual generation is inverted with another generative, yet opposite, visual-to-semantic stream \cite{Schonfeld_2019_CVPR,shen2020invertible}.

Differently to all these methods, our GAN-based architecture is different in the way it synthesizes class embeddings for the unknown classes. Please note that two recent solutions applied a similar idea for the sake of learning a better noise for the GAN \cite{Zhu:ICCV19} 
\cite{marmoreo2021transductive}, but, \textit{to the best of our knowledge, we are the first to synthesize class embeddings.} As a concurrent work to ours, \cite{mancini2021open} seems to approach the open-world scenario as well: but, rather than building upon the ``standard'' (G)ZSL protocol used in computer vision \cite{xian2018tPAMI}, it approaches the ``compositional setup''. That is, seen classes are defined as combinations of tags (\eg, ``wet dog'' or ``furry cat'') and inference has to be done on unknown combinations (\eg, ``furry dog''). Differently to \cite{mancini2021open}, we put no prior on the classes we need to generalize onto (unseen and unknown mainly) as we tackle the challenging generalization gap that requires us, for example, to reject unknown dolphins while not forgetting how to classify seen humpback whales and unseen blue whales.

\paragraph{Rejecting Unknown Categories.}
After the initial formalization of \cite{Scheirer2012} on how to learn in the open world paradigm, many approaches have proposed for letting traditional machine learning models to deal with the \textit{unknown} \cite{Boult2014,Bendale2014, Jain2014MulticlassOS, Cevikalp2013, Cevikalp2016, Scherreik2016, Cevikalp2017,Cevikalp2017b,Zhang2016,Pedro2017,Rudd2015,Vignotto2018,Fei2016,Vareto2017,Manuel2018}. The interested reader may refer to \cite{Geng2020RecentAI} for an overview. 

Leveraging the widespread usage of softmax classifier as the default classifier of deep neural networks, Openmax \cite{OpenMax}, proposed a meta-learning algorithm so that the probability of a data point to be an outlier can be modelled generating an extra-bin which estimate the probability of rejecting the given instance when recognized as outlier. Since then, a few algorithmic variants have been applied to Openmax, ranging from the usage of data-driven preconditioning \cite{GeGenerative2017} 
to conterfactual learning \cite{Neal_2018_ECCV}. In our case, we do not change Openmax in its algorithmic implementation, but, rather, we fed it by data which are ``much more difficult'' to manage 
as compared to prior art. In fact, we ask Openmax not only to recognize seen classes, but also two different types of categories for which visual data are not available (unseen and unknown). Prior art in Openmax only considers seen vs. unknown \cite{OpenMax} or seen vs. unseen \cite{gune2019generalized} and, to the best of our knowledge, we are the first to jointly consider seen, unseen and unknown. 
\section{Open Zero-Shot Learning}\label{sez:OZSL}

In this Section, we relax the closed-world assumption that constraints GZSL methods in knowing class embeddings for all categories (both seen $\mathcal{S}$ and unseen ones $\mathcal{U}$): we therefore attempt to reject unknown categories while not forgetting seen and unseen ones. 
We do so by proposing OZSL, in which we augment $\mathcal{S}$ and $\mathcal{U}$ with a third set of classes, dubbed \textit{unknown}, and denoted by $\Omega$. Unknown classes are deprived of both visual data and class embeddings (see Fig. \ref{fig:OZSL}). We formalize the OZSL problem by instantiating evaluation protocols, datasets and error metrics. We root these in GZSL to ease the transfer of the zero-shot learning community towards the new OZSL paradigm. 

\paragraph{OZSL evaluation protocol.} In GZSL, seen classes $\mathcal{S}$ are provided of data which are triplets $[\mathbf{x},y,\mathcal{C}_y]$: $\mathbf{x}$ are vectorial visual embeddings extracted from a deep convnet (usually, ResNet101 \cite{xian2018tPAMI}) fed by related images, $y$ is the class label and $\mathcal{C}_y$ is a class embeddings (\eg, a list of manually-defined attributes describing the class that are converted into float numbers ranged in $[0,1]$ through Osherson's default probability scores \cite{lampert2009learning}). Unseen classes $\mathcal{U}$ are instead only given of class embeddings (and labels) $[y,\mathcal{C}_y]$ at training time, hence totally missing visual data. 

In OZSL, together with the recognition of seen and unseen classes, we encompass potentially infinitely many classes at inference time. In fact, in addition to classify examples from $\mathcal{S}$ and $\mathcal{U}$, we also consider examples to be rejected since belonging to \textit{unknown} categories we never observed before (no visual data available) and without class embeddings disclosed to the learner. Thus, unknown classes, denoted by $\Omega$, are totally deprived of any visual or semantic information.

Therefore, the task is to train a zero-shot learner to handle the open-world scenario where, not only it has to recognize 
any unobserved test instance for which visual patterns are apparently matching semantic information of class embeddings, but it has also to avoid to take any decision on instances that seem to have a visual content that is not compatible with any prior semantic knowledge encapsulated in seen and unseen class embeddings.

\paragraph{OSZL datasets.} In order to allow practitioners to provide experimental results in both the closed-world, \ie, GZSL, and the open-world, the proposed OZSL, we build OZSL benchmark datasets rearranging GZSL ones. Specifically, we consider Animals with Attributes (AWA) \cite{AwithA}, Caltech-UCSD Birds 200 2011 (CUB) \cite{CUB}, Scene Understanding (SUN) \cite{SUN}, and Oxford Flowers 102 (FLO) \cite{FLO} since they are, by far, ubiquitous in GZSL literature \cite{Arora2018GeneralizedZL,felix2018multi,huang2019generative,Li:CVPR19,Xian_2019_CVPR,Schonfeld_2019_CVPR,Zhu:ICCV19,Xian_2018_CVPR,Sariyildiz2019GradientMG,Gao2020ZeroVAEGANGU}. We leverage the ``Proposed Splits'' \cite{xian2018tPAMI} to be still enabled to use ImageNet pre-trained models to obtain visual descriptors (which are actually already pre-computed from a ResNet-101 and shared by the authors of \cite{xian2018tPAMI}) and we stick to their proposed subdivision into seen and unseen classes. 
We select unknown categories by sampling from unseen classes. In short, we propose to sample 50\% of the unseen classes of \cite{xian2018tPAMI} and transform them to unknown classes, keeping the remaining 50\% as unseen categories in OZSL. A complete list of seen, unseen and unknown classes for the selected four benchmark datasets is available in the Appendix.

\paragraph{Error metrics.} In GZSL, the performance is usually \cite{xian2018tPAMI} evaluated using the harmonic mean  
\begin{equation}\label{eq:H_GZSL}
    H_{\rm GZSL}=\frac{2R_{\mathcal{S}}R_{\mathcal{U}}}{R_{\mathcal{S}}+R_{\mathcal{U}}},
\end{equation}
between each per-class accuracy $R_{\mathcal{S}}$ and $R_{\mathcal{U}}$, computed over seen and unseen classes, respectively. $R_{\mathcal{S}}$ and $R_{\mathcal{U}}$ are defined as:

\begin{align}
    R_{\mathcal{S}} &=\frac{1}{|\mathcal{S}|}\sum_{s\in\mathcal{S}}R_s=\frac{1}{|\mathcal{S}|}\sum_{s\in\mathcal{S}}\frac{TP_s}{TP_s+FN_s}, \label{eq:R_S} \\
    R_{\mathcal{U}} &=\frac{1}{|\mathcal{U}|}\sum_{u\in\mathcal{U}}R_u=\frac{1}{|\mathcal{U}|}\sum_{u\in\mathcal{U}}\frac{TP_u}{TP_u+FN_u} \label{eq:R_U}.
\end{align}
where, in Eq. \eqref{eq:R_S}, we compute $R_s$, for the fixed seen class $s \in \mathcal{S}$, as the ratio between true positives $TP_s$ and the total test examples of the class $s$, that is the sum of $TP_s$ and the false negatives $FN_s$ for that class. To obtain $R_{\mathcal{S}}$ from $R_s$, $s \in \mathcal{S}$, we average $R_s$ over the whole list of seen classes (having cardinality  $|\mathcal{S}|$). 
Analogous operations are carried out in Eq. \eqref{eq:R_U} to compute $R_{\mathcal{U}}$, but applied to unseen classes in $\mathcal{U}$, instead. The metrics $H_{\rm GZSL}$, $R_{\mathcal{S}}$ and $R_{\mathcal{U}}$ were proposed in \cite{xian2018tPAMI} and adopted by state-of-the-art methods for their experimental validation \cite{mishra2018generative,Arora2018GeneralizedZL,felix2018multi,huang2019generative,Li:CVPR19,Xian_2019_CVPR,Schonfeld_2019_CVPR,Zhu:ICCV19,Xian_2018_CVPR,Sariyildiz2019GradientMG,Gao2020ZeroVAEGANGU}.

In GZSL, given that both seen and unseen classes have to be reliably classified, it makes sense to have error metrics depending upon true positives and false negatives which are computed independently over seen and unseen classes and (harmonically) averaged in order to balance performance over these two sets of categories \cite{xian2018tPAMI}. 

In OZSL, in order to break the closed-world assumption, 
we need to take into account also false positives $FP$. In fact, $FP$ simulates cases where examples are predicted as if they belong to that class, albeit their actual ground-truth class is different. Please note that, since we cannot write explicit multi-class classification accuracy scores for the unknown classes $\Omega$ - since we do not have anything describing them - we have to rely on false positives, for both seen and unseen classes ($FP_s$, for every $s \in \mathcal{S}$, and $FP_u$, for every $u \in \mathcal{U}$), in order to indirectly control the rejection performance. In other words, in order to quantitatively measure the performance of a predictor of seen and unseen classes $\mathcal{S}$ and $\mathcal{U}$, which is also a rejector of unknown classes $\Omega$, we need to control $FP_s$ and $FP_u$, for every $s \in \mathcal{S}$ and $u \in \mathcal{U}$. This will reduce the possibility of wrongly associating generic unknown instances to any of the seen/unseen classes.

Obviously, the prior control on seen/unseen false positives has to be paired with penalization of ``traditional'' mis-classifications in a GZSL sense, since we do not want to gain in robustness towards unknown categories while forgetting how to predict seen or unseen classes. Therefore, we propose to measure performance in OZSL through the harmonic mean
\begin{equation}\label{eq:H_OZSL}
    \boxed{\mathcal{H}_{\rm OZSL}=\frac{2{F1}_{\mathcal{S}}{F1}_{\mathcal{U}}}{{F1}_{\mathcal{S}}+{F1}_{\mathcal{U}}}}
\end{equation}
of the $F1$ scores ${F1}_{\mathcal{S}}$ and ${F1}_{\mathcal{U}}$, over seen and unseen classes, defined as 
\begin{align}
    {F1}_\mathcal{S} &=\frac{1}{|\mathcal{S}|}\sum_{s\in\mathcal{S}}F1_s=\frac{1}{|\mathcal{S}|}\sum_{s\in\mathcal{S}}\frac{2R_sP_s}{R_s+P_s},
\label{eq:F_S} \\
    {F1}_\mathcal{U}&=\frac{1}{|\mathcal{U}|}\sum_{u\in\mathcal{U}}F1_u=\frac{1}{|\mathcal{U}|}\sum_{u\in\mathcal{U}}\frac{2R_uP_u}{R_u+P_u}.
    \label{eq:F_U}
\end{align}
In Eq. \eqref{eq:F_S}, for each seen class $s \in \mathcal{S}$, we compute the harmonic mean $F1_s$ of $R_s$, defined as in Eq. \eqref{eq:R_S}, and the precision $P_s$ relative to $s$. We have that $P_s = \frac{TP_s}{TP_s + FP_s}$, being defined as the ratio of the true positives $TP_s$ for that class and the total test examples classified as belonging to that class, that is the sum of $TP_s$ and false positives $FP_s$. We repeat the analogous operations over unseen classes to obtain $F1_{\mathcal{U}}$, as in Eq. \eqref{eq:F_U}.

We claim that $\mathcal{H}_{\rm OZSL}$, as defined in Eq. \eqref{eq:H_OZSL} extends the prior metric $H_{\rm GZSL}$ (in Eq. \eqref{eq:H_GZSL}) by preserving its property of evaluating a correct classification of seen and unseen categories. Concurrently, with $\mathcal{H}_{\rm OZSL}$, we also inject false positives, formalizing their addition using $F1$ scores, for the sake of controlling any misclassification involving unknown classes: this is a computable proxy to evaluate performance on unknown classes. 

\section{Generating The Unknown}\label{sez:method}

\begin{figure}[t!]
    \centering
    \begin{overpic}[width=\columnwidth]{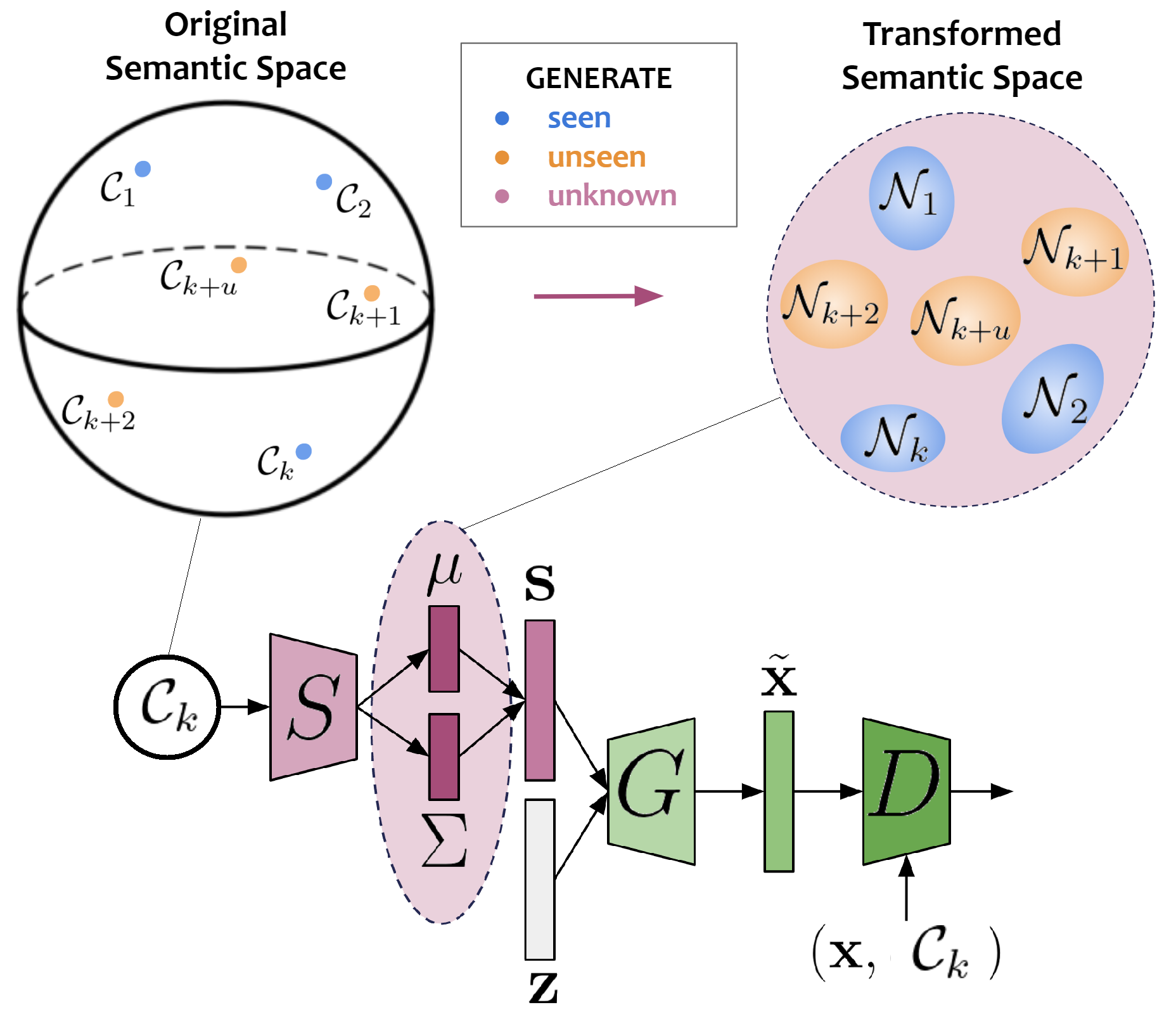}
    \put (87,18) {\scalebox{1.1}{$\mathcal{L}$}}
    \put (87,13) {\scalebox{1.1}{$\mathcal{R}$}}
    \put (87,23) {\scalebox{1.1}{$C$}}
    \end{overpic}
    \caption{Using VAcWGAN, we generate unknown class embeddings (in a transformed semantic space) from which, in turn synthetic unknown visual features can be generated.} 
    \label{fig:SBUG}
\end{figure}

\paragraph{Motivation.} Feature generators for GZSL, such as  \cite{Xian_2018_CVPR} or \cite{narayan20}, leverage the operative
assumption of knowing the class embeddings even for the categories which are unseen at training time. Class embeddings are, in fact, adopted as conditioning factors inside GAN- \cite{Xian_2018_CVPR,Han_2021_CVPR}, VAE- \cite{mishra2018generative,Arora2018GeneralizedZL} or GAN+VAE-based methods \cite{narayan20,Xian_2019_CVPR} to synthesize visual descriptors for the unseen classes. We cannot repeat the very same operation for unknown classes $\Omega$ since we have no class embeddings, but we still need to generate visual features because we do not have them as well. 

\subsection{Direct Unknown Generation (DUG)}\label{sez:DUG}
Exploiting the generative methods \cite{Xian_2018_CVPR,narayan20,Han_2021_CVPR}, is possible to train a generative method, trained only on seen categories, to be conditioned on semantic embeddings to generate corresponding visual features. Thus, once the training is complete, is possible to condition on the class embeddings of the unseen classes to generate unseen visual features. Once both the seen and unseen visual features are available, inspired by \cite{zhang2018mixup}, we take advantage of the MixUp approach to directly generate visual features for the unknown categories.

That is, given two visual features $\mathbf{x}_1$ and $\mathbf{x}_2$, representative of different classes, we mix them with
\begin{equation}
    \mathbf{x}_k=\lambda\mathbf{x}_1+(1-\lambda)\mathbf{x}_2,
\end{equation}
where $\lambda\in[0,1]$ is sampled from a distribution Beta$(\alpha,\beta)$, and the unknown label is assigned to $\mathbf{x}_k$.

The mixed features present mixed traits of the seen and unseen categories, belonging to any of them, and lie in regions of the feature space in between different classes. By labeling them as unknown, we can heuristically build borders for the classification regions for the seen and unseen classes and create a prior knowledge for classifying the unknown features.

\subsection{Semantic Based Unknown Generation (SBUG)}\label{sez:SBUG}
Instead of directly using the visual feature space to generate the unknown features, a different approach that we investigate is to take advantage of the semantic embeddings to generate them.
To this end, we propose to adopt a generative process to learn the distribution of the semantic space, as to learn the region of influence of seen and unseen class embeddings (blue and yellow balls in Fig. \ref{fig:SBUG}). So doing, we can map class embeddings into a transformed semantic space, and we claim that, inside it, we can generate class embeddings for the unknown classes by performing a
 mixing approach similar to the one presented in Section \ref{sez:DUG}
Specifically, we sample the transformed semantic space
``in between'' the region of interest of seen and unseen classes, obtaining synthetic unknown class embeddings. Using them, we generate unknown visual features which help a classifier in rejecting unknown classes while still reliably classifying seen and unseen ones (from real seen and synthetic unseen visual features, respectively).

Thus, differently from DUG, where we can apply the unknown feature generation over an existing methodology, with SBUG we perform an end-to-end training together with the generative process to learn the mapping of the semantic embeddings in a new, more controllable, semantic space. 

\paragraph{A generative process on class embeddings: VAcWGAN.} We introduce a semantic sampler $S$ which is responsible of learning first and second order statistics ($\mu$ and $\Sigma$) for each of the classes $y$ whose semantic embedding is given (seen and unseen). Once trained, we sample a vector $\mathbf{s}$ from a Gaussian distribution of mean $\mu$ and covariance matrix $\Sigma\Sigma^\top$. The role of $S$ is to transform the semantic space through a generative process, as the result of which, seen class embeddings $\mathcal{C}_1,\mathcal{C}_2,\dots,\mathcal{C}_k$, and unseen ones $\mathcal{C}_{k+1},\mathcal{C}_{k+2},\dots,\mathcal{C}_{k+u}$ are mapped into regions of influence. That is, they are mapped into $\mathcal{N}_1,\mathcal{N}_2,\dots,\mathcal{N}_k$ (light blue balls in Fig. \ref{fig:SBUG}) and $\mathcal{N}_{k+1},\mathcal{N}_{k+2},\dots,\mathcal{N}_{k+u}$ (yellow balls in Fig. \ref{fig:SBUG}). We model $\mathcal{N}_1,\mathcal{N}_2,\dots,\mathcal{N}_k,\mathcal{N}_{k+1},\mathcal{N}_{k+2},\dots,\mathcal{N}_{k+u}$ as Gaussian distributions and we use them to sample the conditioning factor $\mathbf{s}$ which, paired to a random noise vector $\mathbf{z}$ is passed to a Wasserstein GAN. This GAN is trained to generate synthetic visual features $\tilde{\mathbf{x}}$ by making them indistinguishable from the real seen features $\mathbf{x}$ extracted by an ImageNet pre-trained ResNet-101 model.

We call the aforementioned architecture variationally-conditioned Wasserstein GAN (VAcWGAN), which is built over the following optimization: $\displaystyle \min_{G,S} \max_D \mathcal{L}$,
where 
\begin{align}\nonumber
    \mathcal{L}(\mathbf{x},\widetilde{\mathbf{x}},\mathbf{s}) & = L^{\rm real}(\mathbf{x},\mathbf{s}) - L^{\rm fake}(\widetilde{\mathbf{x}},\mathbf{s}) \\ &= \mathbb{E}_{\mathbf{x} \sim {\rm real}} \big[ D ( \mathbf{x}, \mathbf{s}) \big] - \mathbb{E}_{\widetilde{\mathbf{x}} \sim {\rm gen}} \big[ D ( \widetilde{\mathbf{x}}, \mathbf{s}) \big]. \label{eq:LLL}
\end{align}
In Eq. \eqref{eq:LLL}, $\mathcal{L}(\mathbf{x},\widetilde{\mathbf{x}},\mathbf{s})$ attempts to align the Wasserstein (Earth Mover) distance \cite{Arjovsky2017WassersteinG} between the distributions of synthesized features $\widetilde{\mathbf{x}}$ over the distribution of the real ones $\mathbf{x}$. We introduce two auxiliary losses for VAcWGAN by jointly considering a standard gradient penalty term \cite{improvedWGAN}
\begin{equation}\nonumber
    \mathcal{R}(\mathbf{x},\widetilde{\mathbf{x}},\mathbf{s}) = \mathbb{E}_{t \in [0,1]}  \big[ (\|\nabla D (t\mathbf{x} + (1-t)\widetilde{\mathbf{x}}, \mathbf{s}) \|_2 - 1 )^2 \big]
\end{equation}
which is commonly acknowledged to regularize the whole generation process, increasing computational stability \cite{improvedWGAN}. We used a cross-entropy classification loss \cite{Xian_2018_CVPR}
\begin{equation}\label{eq:class}
    C(\widetilde{\mathbf{x}}) = - \mathbb{E}_{\widetilde{x} \sim {\rm gen}} \big[ \log p(y|\widetilde{\mathbf{x}}) \big]
\end{equation}

which constraints the softmax probability $p$ of classifying $\widetilde{\mathbf{x}}$ to belong to the class $y$: it has to match the prediction done on $\widetilde{\mathbf{x}}$ when generated from the class embedding $\mathcal{C}_y$ relative to the class $y$. 

The pseudocode to train VAcWGAN is provided in Alg. \ref{alg:VAWGAN}, while additional implementation details are available in the Appendix. 

\begin{algorithm}[t!]
    \SetAlgoLined
    Randomly initialize $S,G$ and $D$ \;
    Generate $\widetilde{\mathbf{x}}$ and pre-train the softmax classifier $p$
    \While{not converged}{
        \For{$i \gets 1$ \textbf{to} $M$}{
            update $D$ using $\mathcal{L}$ and $\mathcal{R}$
        }
        Synthesized unseen features $\widetilde{\mathbf{x}}$ \;
        Update $G$ using $L^{\rm fake}$, $\mathcal{R}$ and $C$\;
            Update $S$ using $L^{\rm fake}$, $\mathcal{R}$ and $C$\;
    }
\caption{Training ${\rm VAcWGAN}$}
\label{alg:VAWGAN}
\end{algorithm}

\paragraph{Semantic Based Unknown Generation.}
We train VAcWGAN using \textit{seen data only}. In addition to generating unseen visual features (as commonly done in GZSL, see Section \ref{sez:rl}), we can also generate the unknown with a two-stages process. Given the generative process that VAcWGAN endow on class embeddings, we estimate the region of interest $\mathcal{N}_1 \cup \mathcal{N}_2\cup  \dots \cup \mathcal{N}_k\cup\mathcal{N}_{k+1}\cup\mathcal{N}_{k+2}\cup \dots \cup\mathcal{N}_{k+u}$ of both seen and unseen classes (in a transformed semantic space). We can exploit the complementary of it (\ie, the pink region in Figure \ref{fig:SBUG}) to sample class embeddings that
lie in the new semantic space in the regions in between the seen and unseen classes by mixing samples of seen and unseen class embeddings.

Specifically we sample to semantic embeddings for two different classes $\mathcal{C}_i$ and $\mathcal{C}_j$, sample accordingly to the regions of interest $\mathbf{s}_i\sim\mathcal{N}_{i}$ and $\mathbf{s}_j\sim\mathcal{N}_{j}$, and than we mix them with
\begin{equation}
    \mathbf{s}_k=\lambda\mathbf{s}_i+(1-\lambda)\mathbf{s}_j,
\end{equation}
where $\lambda\in[0,1]$ is sampled from a distribution Beta$(\alpha,\beta)$, and the unknown label is assigned to $\mathbf{s}_k$.
Once unknown class embeddings are sampled, they can be used as a conditioning factor to generate visual features that can be ascribed to the unknown classes.

\section{Experiments}\label{sez:exp}

\sidecaptionvpos{figure*}{c}
\begin{SCfigure*}[][h]
    \centering
    \includegraphics[width=0.55\textwidth]{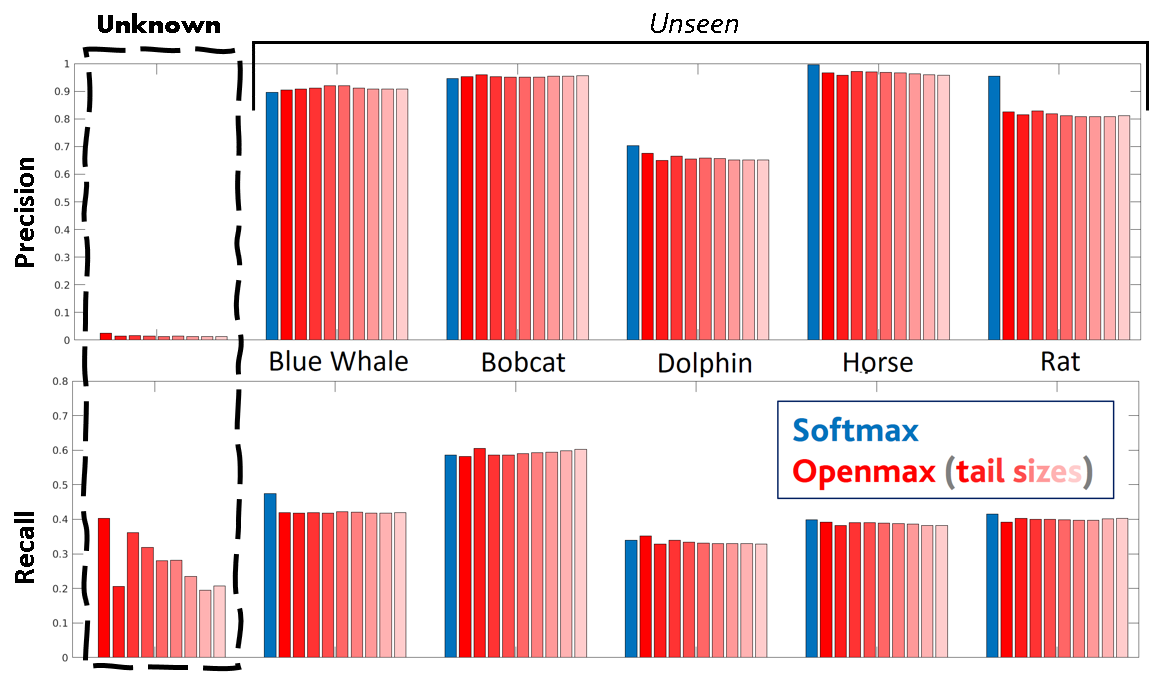}
    \caption{Precision and Recall of CLSWGAN \cite{Xian_2018_CVPR} combined with either Softmax or Openmax \cite{OpenMax} on the AWA dataset. Openmax has been tuned using different tail sizes (ranging from 2 - darkest red - to 10 - lighter red). When adopting state-of-the-art solutions (like \cite{OpenMax}) to cope with the unknown, we argue that the joint presence of unseen classes (which we do not have to forget) prevents Openmax to reliably rejecting the unknown - as it appears to be able to if we remove unseen classes (see \cite{OpenMax}). We perceive this as evidence of the challenges related to OZSL.}
    \label{fig:barsAWA}
\end{SCfigure*}

\begin{table*}[t!]
\centering
\resizebox{0.8\textwidth}{!}{\begin{tabular}{lllll|ll|ll|l}
\hline
\multicolumn{2}{c}{{tf-VAEGAN} \cite{narayan20}} & $P_\Omega$       &    $R_\Omega$     & $F1_{\Omega}$ & \textcolor{Green}{bobcat}     & \textcolor{Green}{giraffe} & \textcolor{Red}{horse}   & \textcolor{Red}{sheep}   & $F1_{\mathcal{U}}$ \\
 \multirow{2}{*}{\textit{AWA}}                  & {Softmax} & 0.00\%    & 0.00\%    & \textit{0.00\%}     & 84.90\%    & 87.61\% & 39.40\% & 59.80\% & \textit{72.22\%} \\
                & {Openmax} & 18.81\%   & 58.24\%   & \textit{28.43\%}    & 92.66\%    & 87.56\% & 37.38\% & 48.94\% & \textit{76.04\%} \\ \hline
\multicolumn{2}{c}{{CLSWGAN} \cite{Xian_2018_CVPR}} & $P_\Omega$       &    $R_\Omega$     & $F1_{\Omega}$ & \textcolor{Green}{blue whale} & \textcolor{Green}{bobcat}  & \textcolor{Red}{dolphin} & \textcolor{Red}{rat}     &  $F1_{\mathcal{U}}$       \\
 \multirow{2}{*}{\textit{AWA}}                  & {Softmax} & 0.00\%    & 0.00\%    & \textit{0.00\%}     & 72.07\%    & 72.36\% & 45.73\% & 56.97\% & \textit{70.42\%} \\
                & {Openmax} & 22.45\%    & 70.87\%   & \textit{34.10\%}     & 77.35\%    & 75.26\% & 46.25\% & 55.75\% & \textit{74.90\%} \\\hline
\end{tabular}}\vspace{2pt}\vspace{2pt}
\resizebox{0.9\textwidth}{!}{\begin{tabular}{lllll|ll|ll|l} \hline
\multicolumn{2}{c}{{tf-VAEGAN} \cite{narayan20}} & $P_\Omega$       &    $R_\Omega$     & $F1_{\Omega}$ & \textcolor{Green}{green violetear}   & \textcolor{Green}{scarlet tanager}           & \textcolor{Red}{tree sparrow}                   & \textcolor{Red}{yellowthroat}     & $F1_{\mathcal{U}}$        \\ 
 \multirow{2}{*}{\textit{CUB}}                 & {Softmax} & 0.00\% & 0.00\%  & \textit{0.00\%} & 89.55\%           & 88.06\%                   & 26.67\%                        & 51.67\%                 & \textit{67.15\% }    \\
                & {Openmax} & 0.80\% & 40.00\% & \textit{1.57\%} & 100.00\%          & 98.33\%                   & 12.31\%                        & 18.90\%                 & \textit{69.43\%}     \\ \hline
\multicolumn{2}{c}{{CLSWGAN} \cite{Xian_2018_CVPR}} & $P_\Omega$       &    $R_\Omega$     & $F1_{\Omega}$ & \textcolor{Green}{bl. cormorant} & \textcolor{Green}{red c woodp} & \textcolor{Red}{orange warb} & \textcolor{Red}{mockingbird} & $F1_{\mathcal{U}}$ \\
  \multirow{2}{*}{\textit{CUB}}                & {Softmax} & 0.00\% & 0.00\%  & \textit{0.00\%} & 83.05\%           & 96.55\%                   & 31.67\%                        & 13.51\%                 & \textit{64.45\%}     \\
                & {Openmax} & 3.35\% & 70.16\% & \textit{6.40\%} & 69.49\%           & 89.66\%                   & 26.39\%                        & 17.22\%                 & \textit{68.53\%}  \\ \hline
\end{tabular}}\vspace{2pt}\vspace{2pt}
\resizebox{0.9\textwidth}{!}{\begin{tabular}{lllll|ll|ll|l} \hline
\multicolumn{2}{c}{{tf-VAEGAN} \cite{narayan20}} & $P_\Omega$       &    $R_\Omega$     & $F1_{\Omega}$     & \textcolor{Green}{purple cone} & \textcolor{Green}{tigerlily} & \textcolor{Red}{pink prim} & \textcolor{Red}{sweetpea} & $F1_{\mathcal{U}}$    \\
\multirow{2}{*}{\textit{FLO}}                & {Softmax} & 0.00\%  & 0.00\%  & \textit{0.00\%}  & 80.65\%            & 93.33\%     & 45.00\%        & 42.86\%    & \textit{65.14\%} \\
                & {Openmax} & 10.16\% & 65.17\% & \textit{17.58\%} & 88.24\%            & 73.68\%     & 37.50\%        & 20.51\%    & \textit{69.78\%} \\ \hline
\multicolumn{2}{c}{{CLSWGAN} \cite{Xian_2018_CVPR}} & $P_\Omega$       &    $R_\Omega$     & $F1_{\Omega}$ & \textcolor{Green}{purple cone} & \textcolor{Green}{camellia}    & \textcolor{Red}{buttercup}    & \textcolor{Red}{azalea}     & $F1_{\mathcal{U}}$   \\ 
\multirow{2}{*}{\textit{FLO}} & {Softmax} & 0.00\%  & 0.00\%  & \textit{0.00\%}  & 88.24\%    & 82.14\%    & 30.90\%       & 46.43\% & \textit{52.56\%} \\
                & {Openmax} & 18.32\% & 81.36\% & \textit{29.91\%} & 80.65\% & 80.36\%    & 32.92\%       & 44.64\%   & \textit{53.80\%} \\ \hline
\end{tabular}}\vspace{2pt}\vspace{2pt}
\resizebox{0.9\textwidth}{!}{
\begin{tabular}{lllll|ll|ll|l} \hline
\multicolumn{2}{c}{{tf-VAEGAN} \cite{narayan20}} & $P_\Omega$       &    $R_\Omega$     & $F1_{\Omega}$  & \textcolor{Green}{hoodoo}                    & \textcolor{Green}{fishpond}       & \textcolor{Red}{bow wind. ind.} & \textcolor{Red}{elevator} & $F1_{\mathcal{U}}$    \\
\multirow{2}{*}{\textit{SUN}}                & {Softmax} & 0.00\% & 0.00\%  & \textit{0.00\%}  & 85.00\%                  & 85.00\%        & 50.00\%             & 35.00\%            & \textit{56.33\%} \\
                & {Openmax} & 2.06\% & 43.35\% & \textit{3.92\%}  & 95.00\%                  & 85.00\%        & 29.41\%             & 20.00\%            & \textit{61.68\%} \\ \hline
\multicolumn{2}{c}{{CLSWGAN} \cite{Xian_2018_CVPR}} &  $P_\Omega$       &    $R_\Omega$     & $F1_{\Omega}$ & \textcolor{Green}{car seat} & \textcolor{Green}{church indoor} & \textcolor{Red}{field cult.}   & \textcolor{Red}{ballroom} & $F1_{\mathcal{U}}$    \\
\multirow{2}{*}{\textit{SUN}} & {Softmax} & 0.00\% & 0.00\%  & \textit{0.00\%}  & 86.36\%                  & 70.59\%        & 22.67\%             & 22.92\%            & \textit{50.44\%} \\
                & {Openmax} & 6.94\% & 58.99\% & \textit{12.43\%} & 94.44\%                  & 75.00\%        & 24.14\%             & 38.89\%            & \textit{53.76\%} \\ \hline
\end{tabular}}
\caption{Baseline methods for OZSL evaluated on their capability of rejecting unknown (treated as a separated class for which precision $P_\Omega$, recall $R_\Omega$, and F1 $F1_\Omega$ scores can be computed. 
We also focus here on classifying unseen classes, reporting the average F1 score $F1_{\mathcal{U}}$ over them, while also reporting the per-class F1 score for two exemplar classes whose performance is above the mean (6th, 7th columns, marked in green), and two classes that are below it (8th, 9th columns, marked in red). 
We observe that, generically, Openmax achieves high recall and low precision. The softmax is not capable of rejecting, therefore $P_\Omega = R_\Omega = F1_\Omega =$~0\%.}
\label{tab:baseline}
\end{table*}

\begin{table*}[]
\begin{center}
    
\resizebox{\textwidth}{!}{%
\begin{tabular}{l|l|l|l|l|l|l|l|l|l|l|l|l|}
 & \multicolumn{3}{c|}{AWA} & \multicolumn{3}{c|}{CUB} & \multicolumn{3}{c|}{FLO} & \multicolumn{3}{c|}{SUN} \\
 & \multicolumn{1}{c|}{$F1_{\mathcal{U}}$} & \multicolumn{1}{c|}{$F1_{\mathcal{S}}$ } & \multicolumn{1}{c|}{$\mathcal{H}_{\rm OZSL}$} & \multicolumn{1}{c|}{$F1_{\mathcal{U}}$} & \multicolumn{1}{c|}{$F1_{\mathcal{S}}$} & \multicolumn{1}{c|}{$\mathcal{H}_{\rm OZSL}$} & \multicolumn{1}{c|}{$F1_{\mathcal{U}}$} & \multicolumn{1}{c|}{$F1_{\mathcal{S}}$} & \multicolumn{1}{c|}{$\mathcal{H}_{\rm OZSL}$} & \multicolumn{1}{c|}{$F1_{\mathcal{U}}$} & \multicolumn{1}{c|}{$F1_{\mathcal{S}}$} & \multicolumn{1}{c|}{$\mathcal{H}_{\rm OZSL}$} \\ \hline
\multicolumn{1}{|l|}{VAcWGAN} & 
50.31\% & 
64.84\% & 
56.66\% & 
45.08\% & 	
49.23\% &  
47.34\% &  
	\textbf{47.68\%} &
	{72.69\%} &
	\textbf{57.59\%} &

\textbf{38.05\%} &  		
{37.33\%} & 
{37.68\%} \\ \hline 
\multicolumn{1}{|l|}{VAcWGAN + DUG} & 
\textbf{51.83\%} & 
{65.18\%} & 
{57.74\%} & 
{45.48\%} & 	
{49.98\%} &  
{47.63\%} & 
{46.25\%} &
\textbf{73.30\%} &
{56.71\%} &	
33.87\%	&
\textbf{38.37\%} &
\textbf{37.68}\% \\ 	
\multicolumn{1}{|l|}{VAcWGAN + SBUG} & 
{50.62\%} & 
\textbf{68.28\%} & 
\textbf{58.14\%} & 
{45.59\%} & 	
\textbf{51.42\%}&  
{48.04\%}&  
46.01\%	&  
69.40\%&  
55.33\%&  
{37.65\%} &  		
38.07\% & 
{35.98\%} \\
\multicolumn{1}{|l|}{VAcWGAN + DUG + SBUG} & 
\multicolumn{1}{c|}{51.63\%} &		
\multicolumn{1}{c|}{66.20\% } & 
\multicolumn{1}{c|}{58.01\%} & 
\multicolumn{1}{c|}{\textbf{45.76\%}} & 		\multicolumn{1}{c|}{51.28\%} & 
\multicolumn{1}{c|}{\textbf{48.36\%}} & 
\multicolumn{1}{c|}{47.24\%} & 
\multicolumn{1}{c|}{72.00\%}& 
\multicolumn{1}{c|}{57.05\%} & 
\multicolumn{1}{c|}{34.83\%} & 
\multicolumn{1}{c|}{38.53\%} & 
\multicolumn{1}{c|}{36.59\%} %
\\ \hline 
\end{tabular}
}
\end{center}
\vspace{-10pt}
\caption{Direct and semantic unknown generation (DUG and SBUG) for the WAcWGAN model (check Sec. \ref{sez:SBUG}).}
\label{tab:50_50_2}
\end{table*}

\begin{table*}[]
\begin{center}
    
\resizebox{\textwidth}{!}{%
\begin{tabular}{l|l|l|l|l|l|l|l|l|l|l|l|l|}
 & \multicolumn{3}{c|}{AWA} & \multicolumn{3}{c|}{CUB} & \multicolumn{3}{c|}{FLO} & \multicolumn{3}{c|}{SUN} \\
 & \multicolumn{1}{c|}{$F1_{\mathcal{U}}$} & \multicolumn{1}{c|}{$F1_{\mathcal{S}}$ } & \multicolumn{1}{c|}{$\mathcal{H}_{\rm OZSL}$} & \multicolumn{1}{c|}{$F1_{\mathcal{U}}$} & \multicolumn{1}{c|}{$F1_{\mathcal{S}}$} & \multicolumn{1}{c|}{$\mathcal{H}_{\rm OZSL}$} & \multicolumn{1}{c|}{$F1_{\mathcal{U}}$} & \multicolumn{1}{c|}{$F1_{\mathcal{S}}$} & \multicolumn{1}{c|}{$\mathcal{H}_{\rm OZSL}$} & \multicolumn{1}{c|}{$F1_{\mathcal{U}}$} & \multicolumn{1}{c|}{$F1_{\mathcal{S}}$} & \multicolumn{1}{c|}{$\mathcal{H}_{\rm OZSL}$} \\ \hline
\multicolumn{1}{|l|}{CSLWGAN} & \multicolumn{1}{c|}{52.37\%
} & \multicolumn{1}{c|}{65.52\%
} & \multicolumn{1}{c|}{58.21\% 
} & \multicolumn{1}{c|}{47.01\% 
} & \multicolumn{1}{c|}{52.77\%
} & \multicolumn{1}{c|}{49.73\% 
} & \multicolumn{1}{c|}{48.47\%
} & \multicolumn{1}{c|}{76.08\%
} & \multicolumn{1}{c|}{59.22\% 
} & \multicolumn{1}{c|}{36.31\%
} & \multicolumn{1}{c|}{38.77\%
} & \multicolumn{1}{c|}{37.50\% 
} \\ 

\multicolumn{1}{|l|}{CLSWGAN + DUG} & \textbf{57.34\%}
  & \textbf{67.41\%}
 & \textbf{61.97\%}
 & \textbf{47.76\%}
  & \textbf{53.29\%}
 & \textbf{50.37\%}
& \textbf{49.34\%}
 & \textbf{76.48\%}
  & \textbf{59.98\%}
 & \textbf{36.92\%}
 & \textbf{39.64\%}
 & \textbf{38.23\%}
 \\ \hline
\multicolumn{1}{|l|}{tf-VAEGAN} & 55.49\% 
& 71.47\% 
& 62.48\% 
& \textbf{52.24\%} 
& 56.62\%
& 54.34\% 
& \textbf{54.78\%}
& 80.00\% 
& 65.03\% 
& 43.00\% 
& \textbf{41.09\%} 
& 42.02\%\\ 

\multicolumn{1}{|l|}{tf-VAEGAN + DUG} &  \textbf{60.56\%}
& \textbf{71.73\% }
& \textbf{65.68\%} 
& 51.60\% 
& \textbf{57.92\%} 
& \textbf{54.58\%} 
& 54.35\% 
& \textbf{81.33\%} 
& \textbf{65.15\%} 
& \textbf{46.53\%} 
& 40.76\% 
& \textbf{42.10\%} \\ \hline
\multicolumn{1}{|l|}{CEZSL} & 
51.70\% &
\textbf{71.66\%} &	
60.07\% &5
39.43\%&
56.83\%&	
46.56\%&
39.21\%&
\textbf{85.36\%}&
53.74\%& 
33.35\%&
\textbf{30.83\%}&
32.04\% 
\\ 

\multicolumn{1}{|l|}{CEZSL + DUG} &
\textbf{55.76\%}	 &
71.30\% &
\textbf{62.58\%} & 
\textbf{40.52\%} &	
\textbf{57.54\%} &
\textbf{47.55\%} & 
\textbf{40.26\%}& 
85.25\%&
\textbf{54.69\%}& 
\textbf{35.17\%}	&
30.15\%	&
\textbf{32.47\%}
\\ 
\hline
\end{tabular}
}
\end{center}
\vspace{-10pt}
\caption{The effect of direct unknown generation (DUG) applied on three benchmark methods: CSLWGAN \cite{Xian_2018_CVPR}, tf-VAEGAN \cite{narayan20} and CEZSL \cite{Han_2021_CVPR} on the proposed OZSL splits for AWA \cite{AwithA}, CUB \cite{CUB}, FLO \cite{FLO} and SUN \cite{SUN} datasets.}
\label{tab:50_50}
\end{table*}

In order to understand how difficult OZSL is, in this Section, we inspect the performance achievable by combining state-of-the-art feature generators (\cite{Xian_2018_CVPR,narayan20} for ZSL, combining them with a state-of-the-art classifier for open recognition: Openmax \cite{OpenMax}. 

In Figure \ref{fig:barsAWA}, we compare a standard softmax (in blue) vs. Openmax  (in red), 
using a CLSWGAN \cite{Xian_2018_CVPR} for unseen (but not unknown) feature generation. On average, we do not register a sharp overall advantage of openmax versus softmax (only +0.6\% boost in precision and +0.9\% for recall given by Openmax tuned with tail size 2).
In principle, openmax is theoretically superior to a softmax operator, since it is capable of performing rejection. However, experimentally, we did non register a direct consequence in a superior classification performance. In fact, the recall of Openmax in rejecting the unknown ($R_\Omega = 22.12\%$) is not so dissimilar to the recall values scored on some unseen classes (\eg, Horse or Bobcat in Fig. \ref{fig:barsAWA}). As shown in the literature \cite{OpenMax}, Openmax is arguably a state-of-the-art method to perform rejection, while also recognizing seen classes only. However, a plain transfer of Openmax from its original framework to OZSL (in which, unseen classes have to be recognized \textit{as well}) is not enough to solve the problem. We deem this as  evidence for the intrinsic difficulty of OZSL which appears as arguably hard - and thus intriguing.
  
Even resorting to a better feature generator is not enough to solve the problem, as we show in Table \ref{tab:baseline}. Therein, we provide
a comparison between the $F1_{\Omega}$ score computed over unknown classes, pretending to treat all unknown classes into a macro-container called ``unknown'' (while in principle unknown instances belong to potentially infinite different unknown categories). In addition, we also check $F1_{\mathcal{U}}$, the F1 score over unseen classes only. While exploiting a better model than tf-VAEGAN, we can surely always state 
that Openmax yields a better $F1_{\mathcal{U}}$ with respect to CLSWGAN with Openmax (76.04\% vs. 74.90\% on AWA, 69.43\% vs. 68.53\% on CUB, 69.78\% vs. 53.80\% on FLO and 61.68\% vs. 53.76\% on SUN), while also improving tf-VAEGAN with softmax (improving $F1_{\mathcal{U}}$ by +4\% on AWA, +2\% on CUB, +5\% on FLO and +6\% on SUN). But, this result comes at the price of loosing in $F1_{\Omega}$, whose performance is much higher when using CLSWGAN as opposed to tf-VAEGAN (-6\% on AWA, -4\% on CUB, -12\% on FLO and -8\% on SUN). 

Apparently, taking existing feature generators for ZSL (CLSWGAN \cite{Xian_2018_CVPR} or tf-VAEGAN \cite{narayan20}) and combining them with state-of-the-art methods in open recognition (like openmax \cite{OpenMax}) is not enough to solve OZSL which appears as an intriguing problem. To start solving it, in the next Section, we evaluate the effect of performing unknown feature generation.

\subsection{OZSL through Unknown Generation}

In Table \ref{tab:50_50_2} we perform an experimental evaluation between the two strategies of unknown feature generation we presented: DUG (Direct Unknown Generation, Sec. \ref{sez:DUG}) and SBUG (Semantic Based Unknown Generation, Sec. \ref{sez:SBUG}). We combined DUG and SBUG separately and/or jointly to the VAcWGAN architecture (Sec. \ref{sez:method}).
In the sharp majority of the cases, doing unknown feature generation (with either DUG, SBUG or DUG+SBUG) is better than not doing it. We deem this result highly non-trivial: by attempting to learn how to synthesize unknown descriptors, we are simultaneously better shaping the region of interest of seen and unseen classes, so that the $F1_{\mathcal{U}}$ and $F1_{\mathcal{S}}$ metrics often improve (and so happens for their harmonic mean  $\mathcal{H}_{\rm OZSL}$ as well). For instance, the unknown feature generation improves by +1.48\%, +1.02\% and +0.18\% the performance of WAcWGAN on AWA, CUB and SUN respectively, while considering the  $\mathcal{H}_{\rm OZSL}$ metric and the SBUG, DUG + SBUG and DUG techniques, respectively.

Over DUG and SBUG, the former is advantageous over the latter because it acts directly on the visual space (so that the feature generator has not to be re-trained for unknown synthesis, but can be adapted to it). In view of this consideration, we can apply unknown feature generation to three state-of-the approaches for ZSL, better tailoring them towards the open ZSL regime. Namely, we consider the following methods: the Wasserstein generative adversarial network conditioned on class embeddings (CLSWGAN) \cite{Xian_2018_CVPR} and its extension tf-VAEGAN \cite{narayan20} in which this architecture is paired with a variational auto-encoder to boost the generation stage. We also considered the usage of contrastive learning as adopted in CEZSL \cite{Han_2021_CVPR} in tandem with adversarial training. We combine CLSWGAN, tf-VAEGAN and CEZSL with the direct unknown generation that we presented in Sec. \ref{sez:DUG} and dubbed here ``DUG'' for brevity. As we show in Table \ref{tab:50_50}, the adoption of DUG is always able to improve in performance all the considered baseline approaches with respect to the $\mathcal{H}_{\rm OZSL}$ metric (\eg, +3.76\% on AWA for CLSWGAN, +0.73\% on SUN for CLSWGAN, +0.99\% on CUB for CEZSL and +3.2\% on AWA for tf-VAEGAN). Again, we interpret the consistent improvements that we registered as evidence for the effectiveness of performing unknown feature generation for OZSL. 

\section{Conclusions \& Future Work}\label{sez:conc}

In this paper, we proposed a novel paradigm, called Open Zero-Shot Learning, extending traditional ZSL frameworks towards the additional rejection of unknown categories (neither visually nor semantically described) while still recognizing seen and unseen classes. Using the experimental protocols and error metrics that we proposed, our experimental findings suggest that attempting to synthesize unknown descriptors (to be rejected) seems a viable solution for OZSL.

Future works will be aimed at adopting techniques from out-of-domain generalization to better achieve the way we explore the semantic/visual spaces while seeking better strategies to generate the unknown.

{\small
\bibliographystyle{ieee_fullname}
\bibliography{mybib}

\begin{thebibliography}{10}\itemsep=-1pt

\bibitem{Arjovsky2017WassersteinG}
Mart{\'i}n Arjovsky, Soumith Chintala, and L{\'e}on Bottou.
\newblock {Wasserstein GAN}.
\newblock {\em ArXiv}, abs/1701.07875, 2017.

\bibitem{Arora2018GeneralizedZL}
Gundeep Arora, Vinay~Kumar Verma, Ashish Mishra, and Piyush Rai.
\newblock Generalized zero-shot learning via synthesized examples.
\newblock {\em The IEEE Conference on Computer Vision and Pattern Recognition
  (CVPR)}, 2018.

\bibitem{Bendale2014}
Abhijit Bendale and Terrance Boult.
\newblock Towards open world recognition.
\newblock 12 2014.

\bibitem{OpenMax}
Abhijit Bendale and Terrance~E Boult.
\newblock Towards open set deep networks.
\newblock In {\em Proceedings of the IEEE conference on computer vision and
  pattern recognition}, pages 1563--1572, 2016.

\bibitem{Cevikalp2016}
Hakan Cevikalp.
\newblock Best fitting hyperplanes for classification.
\newblock {\em IEEE Transactions on Pattern Analysis and Machine Intelligence},
  39:1--1, 07 2016.

\bibitem{Cevikalp2017}
Hakan Cevikalp and Bill Triggs.
\newblock Polyhedral conic classifiers for visual object detection and
  classification.
\newblock pages 4114--4122, 07 2017.

\bibitem{Cevikalp2013}
Hakan Cevikalp, B. Triggs, and Vojtěch Franc.
\newblock Face and landmark detection by using cascade of classifiers.
\newblock {\em 10th IEEE International Conference on Automatic Face and Gesture
  Recognition}, pages 1--7, 01 2013.

\bibitem{Cevikalp2017b}
Hakan Cevikalp and Hasan Yavuz.
\newblock Fast and accurate face recognition with image sets.
\newblock pages 1564--1572, 10 2017.

\bibitem{ChaoEmpGZSL}
Wei-Lun Chao, Soravit Changpinyo, Boqing Gong, and Fei Sha.
\newblock An empirical study and analysis of generalized zero-shot learning for
  object recognition in the wild.
\newblock In {\em The Springer European Conference on Computer Vision (ECCV)},
  2016.

\bibitem{Manuel2018}
Manuel Córdova, Pedro Júnior, Anderson Rocha, and Ricardo Torres.
\newblock Data-fusion techniques for open-set recognition problems.
\newblock {\em IEEE Access}, 6:1--1, 04 2018.

\bibitem{Fei2016}
Geli Fei and Bing Liu.
\newblock Breaking the closed world assumption in text classification.
\newblock pages 506--514, 01 2016.

\bibitem{felix2018multi}
Rafael Felix, Vijay~BG Kumar, Ian Reid, and Gustavo Carneiro.
\newblock Multi-modal cycle-consistent generalized zero-shot learning.
\newblock In {\em The European Conference on Computer Vision (ECCV)}, 2018.

\bibitem{Gao2020ZeroVAEGANGU}
Rui Gao, Xingsong Hou, Jie Qin, Jiaxin Chen, Li Liu, Fan Zhu, Zhao Zhang, and
  Ling Shao.
\newblock Zero-vae-gan: Generating unseen features for generalized and
  transductive zero-shot learning.
\newblock {\em IEEE Transactions on Image Processing}, 29:3665--3680, 2020.

\bibitem{GeGenerative2017}
ZongYuan Ge, Sergey Demyanov, Zetao Chen, and Rahil Garnavi.
\newblock Generative openmax for multi-class open set classification.
\newblock 07 2017.

\bibitem{Geng2020RecentAI}
Chuanxing Geng, Sheng-Jun Huang, and S. Chen.
\newblock Recent advances in open set recognition: A survey.
\newblock {\em IEEE transactions on pattern analysis and machine intelligence},
  2020.

\bibitem{improvedWGAN}
Ishaan Gulrajani, Faruk Ahmed, Mart{\'i}n Arjovsky, Vincent Dumoulin, and
  Aaron~C. Courville.
\newblock Improved training of wasserstein gans.
\newblock In {\em NIPS}, 2017.

\bibitem{gune2019generalized}
Omkar Gune, Amit More, Biplab Banerjee, and Subhasis Chaudhuri.
\newblock Generalized zero-shot learning using open set recognition.
\newblock In {\em BMVC}, page 213, 2019.

\bibitem{Han_2021_CVPR}
Zongyan Han, Zhenyong Fu, Shuo Chen, and Jian Yang.
\newblock Contrastive embedding for generalized zero-shot learning.
\newblock In {\em CVPR}, 2021.

\bibitem{zhang2018mixup}
Yann N.~Dauphin Hongyi~Zhang, Moustapha~Cisse and David Lopez-Paz.
\newblock mixup: Beyond empirical risk minimization.
\newblock {\em International Conference on Learning Representations}, 2018.

\bibitem{huang2019generative}
He Huang, Changhu Wang, Philip~S Yu, and Chang-Dong Wang.
\newblock Generative dual adversarial network for generalized zero-shot
  learning.
\newblock In {\em The IEEE Conference on Computer Vision and Pattern
  Recognition (CVPR)}, 2019.

\bibitem{Jain2014MulticlassOS}
Lalit~P. Jain, W. Scheirer, and T. Boult.
\newblock Multi-class open set recognition using probability of inclusion.
\newblock In {\em ECCV}, 2014.

\bibitem{Pedro2017}
Pedro Júnior, Roberto Souza, Rafael Werneck, Bernardo Stein, Daniel Pazinato,
  Waldir Almeida, Otávio Penatti, Ricardo Torres, and Anderson Rocha.
\newblock Nearest neighbors distance ratio open-set classifier.
\newblock {\em Machine Learning}, 106:1--28, 03 2017.

\bibitem{kingma2017adam}
Diederik~P. Kingma and Jimmy Ba.
\newblock Adam: A method for stochastic optimization, 2017.

\bibitem{AwithA}
CH. Lampert, H. Nickisch, and S. Harmeling.
\newblock Learning to detect unseen object classes by between-class attribute
  transfer.
\newblock In {\em The IEEE Conference on Computer Vision and Pattern
  Recognition (CVPR)}, 2009.

\bibitem{lampert2009learning}
Christoph~H Lampert, Hannes Nickisch, and Stefan Harmeling.
\newblock Learning to detect unseen object classes by between-class attribute
  transfer.
\newblock In {\em Computer Vision and Pattern Recognition (CVPR)}. IEEE, 2009.

\bibitem{Li:CVPR19}
Jingling Li, Mengmeng Jing, Ke Lu, Zhengming Ding, Lei Zhu, and Zi Huang.
\newblock Leveraging the invariant side of generative zero-shot learning.
\newblock In {\em The IEEE Conference on Computer Vision and Pattern
  Recognition (CVPR)}, 2019.

\bibitem{mancini2021open}
Massimiliano Mancini, Muhammad~Ferjad Naeem, Yongqin Xian, and Zeynep Akata.
\newblock Open world compositional zero-shot learning.
\newblock In {\em CVPR}, 2021.

\bibitem{marmoreo2021transductive}
Federico Marmoreo, Jacopo Cavazza, and Vittorio Murino.
\newblock Transductive zero-shot learning by decoupled feature generation.
\newblock In {\em Proceedings of the IEEE/CVF Winter Conference on Applications
  of Computer Vision}, pages 3109--3118, 2021.

\bibitem{mikolov2013distributed}
Tomas Mikolov, Ilya Sutskever, Kai Chen, Greg~S Corrado, and Jeff Dean.
\newblock Distributed representations of words and phrases and their
  compositionality.
\newblock In {\em Advances in neural information processing systems}, pages
  3111--3119, 2013.

\bibitem{mishra2018generative}
Ashish Mishra, Shiva Krishna~Reddy, Anurag Mittal, and Hema~A Murthy.
\newblock A generative model for zero shot learning using conditional
  variational autoencoders.
\newblock In {\em The IEEE Conference on Computer Vision and Pattern
  Recognition (CVPR) Workshops}, pages 2188--2196, 2018.

\bibitem{narayan20}
Sanath Narayan, Akshita Gupta, Fahad~Shahbaz Khan, Cees~GM Snoek, and Ling
  Shao.
\newblock Latent embedding feedback and discriminative features for zero-shot
  classification.
\newblock In {\em The European Conference on Computer Vision (ECCV)}, 2020.

\bibitem{Neal_2018_ECCV}
Lawrence Neal, Matthew Olson, Xiaoli Fern, Weng-Keen Wong, and Fuxin Li.
\newblock Open set learning with counterfactual images.
\newblock In {\em Proceedings of the European Conference on Computer Vision
  (ECCV)}, September 2018.

\bibitem{FLO}
M-E. Nilsback and A. Zisserman.
\newblock A visual vocabulary for flower classification.
\newblock In {\em The IEEE Conference on Computer Vision and Pattern
  Recognition (CVPR)}, 2006.

\bibitem{Rudd2015}
Ethan Rudd, Lalit Jain, Walter Scheirer, and Terrance Boult.
\newblock The extreme value machine.
\newblock {\em IEEE Transactions on Pattern Analysis and Machine Intelligence},
  PP, 06 2015.

\bibitem{Sariyildiz2019GradientMG}
Mert~B{\"u}lent Sariyildiz and Ramazan~Gokberk Cinbis.
\newblock Gradient matching generative networks for zero-shot learning.
\newblock {\em 2019 IEEE/CVF Conference on Computer Vision and Pattern
  Recognition (CVPR)}, pages 2163--2173, 2019.

\bibitem{Scheirer2012}
Walter Scheirer, Anderson Rocha, Archana Sapkota, and Terrance Boult.
\newblock Towards open set recognition.
\newblock {\em IEEE transactions on pattern analysis and machine intelligence},
  11 2012.

\bibitem{Boult2014}
W.~J. {Scheirer}, L.~P. {Jain}, and T.~E. {Boult}.
\newblock Probability models for open set recognition.
\newblock {\em IEEE Transactions on Pattern Analysis and Machine Intelligence},
  36(11):2317--2324, 2014.

\bibitem{Scherreik2016}
Matthew Scherreik and Brian Rigling.
\newblock Open set recognition for automatic target classification with
  rejection.
\newblock {\em IEEE Transactions on Aerospace and Electronic Systems},
  52:632--642, 04 2016.

\bibitem{Schonfeld_2019_CVPR}
Edgar Schonfeld, Sayna Ebrahimi, Samarth Sinha, Trevor Darrell, and Zeynep
  Akata.
\newblock Generalized zero- and few-shot learning via aligned variational
  autoencoders.
\newblock In {\em The IEEE Conference on Computer Vision and Pattern
  Recognition (CVPR)}, June 2019.

\bibitem{shen2020invertible}
Yuming Shen, Jie Qin, Lei Huang, Li Liu, Fan Zhu, and Ling Shao.
\newblock Invertible zero-shot recognition flows.
\newblock In {\em European Conference on Computer Vision}, pages 614--631.
  Springer, 2020.

\bibitem{Vareto2017}
Rafael Vareto, Samira Silva, Filipe Costa, and William Schwartz.
\newblock Towards open-set face recognition using hashing functions.
\newblock 10 2017.

\bibitem{Vignotto2018}
Edoardo Vignotto and Sebastian Engelke.
\newblock Extreme value theory for open set classification - gpd and gev
  classifiers, 08 2018.

\bibitem{CUB}
P. Welinder, S. Branson, T. Mita, C. Wah, F. Schroff, S. Belongie, and P.
  Perona.
\newblock {Caltech-UCSD Birds 200}.
\newblock Technical Report CNS-TR-2010-001, California Institute of Technology,
  2010.

\bibitem{xian2018tPAMI}
Yongqin Xian, Christoph~H Lampert, Bernt Schiele, and Zeynep Akata.
\newblock Zero-shot learning-a comprehensive evaluation of the good, the bad
  and the ugly.
\newblock {\em The IEEE Transactions on Pattern Analysis and Machine
  Intelligence}, 2018.

\bibitem{Xian_2018_CVPR}
Yongqin Xian, Tobias Lorenz, Bernt Schiele, and Zeynep Akata.
\newblock Feature generating networks for zero-shot learning.
\newblock In {\em The IEEE Conference on Computer Vision and Pattern
  Recognition (CVPR)}, June 2018.

\bibitem{Xian_2019_CVPR}
Yongqin Xian, Saurabh Sharma, Bernt Schiele, and Zeynep Akata.
\newblock {F-VAEGAN-D2: A Feature Generating Framework for Any-Shot Learning}.
\newblock In {\em The IEEE Conference on Computer Vision and Pattern
  Recognition (CVPR)}, June 2019.

\bibitem{SUN}
J. {Xiao}, J. {Hays}, K.~A. {Ehinger}, A. {Oliva}, and A. {Torralba}.
\newblock Sun database: Large-scale scene recognition from abbey to zoo.
\newblock In {\em The IEEE Conference on Computer Vision and Pattern
  Recognition (CVPR)}, 2010.

\bibitem{Zhang2016}
He Zhang and Vishal Patel.
\newblock Sparse representation-based open set recognition.
\newblock {\em IEEE Transactions on Pattern Analysis and Machine Intelligence},
  PP:1--1, 09 2016.

\bibitem{zhu2018generative}
Yizhe Zhu, Mohamed Elhoseiny, Bingchen Liu, Xi Peng, and Ahmed Elgammal.
\newblock A generative adversarial approach for zero-shot learning from noisy
  texts.
\newblock In {\em Proceedings of the IEEE conference on computer vision and
  pattern recognition}, pages 1004--1013, 2018.

\bibitem{Zhu:ICCV19}
Yizhe Zhu, Jianwen Xie, Bingchen Liu, and Ahmed Elgammal.
\newblock Learning feature-to-feature translator by alternating
  back-propagation for generative zero-shot learning.
\newblock In {\em The IEEE International Conference on Computer Vision (ICCV)},
  2019.

\end{thebibliography}
}

\newpage

\section*{A. VAcWGAN: implementation details}
We implement $G$, $D$ and $S$ as single hidden layer neural networks with hidden layer of size 4096 for $G$ and $D$ and 2048 for $S$ with leaky ReLU activation for all. 
$S$ takes as input the class embeddings $\mathcal{C}$ and gives as output mean and gives as output mean vector $\mu$ and $\log(\sqrt{\sigma})$ of the same size of $\mathcal{C}$. 
$G$ takes as input the vector $\mathbf{s}$, sampled from the Gaussian distribution defined by $\mu$ and $\log(\sqrt{\sigma})$  concatenated with a noise vector $\textbf{z}$ of the same size of $\mathbf{s}$ sampled from a multivariate normal distribution $\mathcal{N}(\textbf{0},\textbf{I})$, where $\textbf{0}$ a vector of zeros and $\textbf{I}$ and identity matrix, and output a visual feature vectors $\mathbf{}$ (of size 2048 and ReLU activation).
$D$ takes as input visual feature vectors with the related class embedding $\mathcal{C}$ and output an unconstrained real number.
To compute the regularization classification loss we directly classify the synthesized visual features $\mathbf{}$ with a pre-trained softmax.
$M$ of Alg. 1 (in the paper) is fixed to 5. Adam \cite{kingma2017adam} is used as optimizer.
\section*{B. Proposed Splits for OZSL}
In this pages, we provide the actual unseen and unknown classes that we considered for AWA \cite{AwithA}, CUB \cite{CUB}, SUN \cite{SUN} and FLO \cite{FLO}. In the following tables,~\cmark~will denote class to be unseen for a given split (representing that the class embedding is disclosed) while~\xmark~denotes those classes for which the class embedding is not available while visual data are missing as well (\ie, the unknown). For brevity, we omit from the following tables the list of seen classes (provided of both visual and semantic data) since this list is overlapping with the seen classes from the ``Proposed Splits'' of the survey \cite{xian2018tPAMI}. We operated this choice to make our proposed OZSL setup complementary to the close-world setup of generalized zero-shot learning, so that practitioners can gradually shift towards the open-world regime - handling possibly many unknown classes while not forgetting neither seen nor unseen ones.

\vspace{10pt}

\begin{supertabular} {|p{.26\textwidth}  H | p{.05\textwidth}  H |} 
\hline
\multicolumn{4}{|c|}{\bf OZSL : AWA \cite{AwithA}} \\
\textit{Class Name} & \textit{20-80} & \textit{unseen} & \textit{80-20} \\ \hline
horse  & \cmark & \cmark & \cmark \\\hline
blue+whale & \cmark & \cmark & \cmark \\\hline
sheep & \xmark & \xmark & \cmark \\\hline
seal & \xmark & \xmark & \cmark \\\hline
bat & \xmark & \xmark & \xmark \\\hline
giraffe & \xmark & \xmark & \xmark \\\hline
rat & \xmark & \cmark & \cmark \\\hline
bobcat & \xmark & \cmark & \cmark \\\hline
walrus & \xmark & \xmark & \cmark \\\hline
dolphin & \xmark & \cmark & \cmark \\\hline

\end{supertabular}
\newpage

\begin{supertabular}{|p{.26\textwidth} H | p{.05\textwidth}  H |} 
\hline
\multicolumn{4}{|c|}{\bf OZSL : CUB \cite{CUB}} \\
\textit{Class Name} & \textit{20-80} & \textit{unseen} & \textit{80-20} \\ \hline
004.Groove\_billed\_Ani & \xmark & \cmark & \cmark \\\hline
012.Yellow\_headed\_Blackbird & \xmark & \cmark & \cmark  \\\hline
023.Brandt\_Cormorant & \cmark & \cmark & \cmark \\\hline
026.Bronzed\_Cowbird & \xmark & \cmark & \cmark  \\\hline
028.Brown\_Creeper & \cmark & \cmark & \cmark  \\\hline
031.Black\_billed\_Cuckoo & \xmark & \xmark & \cmark  \\\hline 
033.Yellow\_billed\_Cuckoo & \xmark & \xmark & \xmark \\\hline
043.Yellow\_bellied\_Flycatcher & \xmark & \xmark & \xmark \\\hline
045.Northern\_Fulmar &  \xmark & \cmark & \cmark  \\\hline
049.Boat\_tailed\_Grackle &  \xmark & \xmark & \cmark  \\\hline 
052.Pied\_billed\_Grebe & \xmark & \xmark & \xmark \\\hline
055.Evening\_Grosbeak &  \xmark & \xmark & \cmark  \\\hline 
070.Green\_Violetear &  \cmark & \cmark & \cmark  \\\hline
072.Pomarine\_Jaeger & \xmark & \xmark & \cmark  \\\hline 
077.Tropical\_Kingbird &  \xmark & \xmark & \cmark  \\\hline 
084.Red\_legged\_Kittiwake &  \xmark & \cmark & \cmark  \\\hline
087.Mallard & \xmark & \cmark & \cmark  \\\hline
091.Mockingbird & \cmark & \cmark & \cmark  \\\hline
094.White\_breasted\_Nuthatch & \xmark & \xmark & \xmark \\\hline
097.Orchard\_Oriole & \xmark & \cmark & \cmark  \\\hline
098.Scott\_Oriole & \xmark & \xmark & \cmark  \\\hline 
103.Sayornis & \xmark & \xmark & \cmark \\\hline 
104.American\_Pipit & \xmark & \xmark & \xmark \\\hline
111.Loggerhead\_Shrike & \xmark & \xmark & \xmark \\\hline
113.Baird\_Sparrow &  \xmark & \cmark & \cmark  \\\hline
119.Field\_Sparrow & \xmark & \cmark & \cmark  \\\hline
123.Henslow\_Sparrow & \xmark & \xmark & \cmark  \\\hline 
124.Le\_Conte\_Sparrow & \xmark & \cmark & \cmark  \\\hline
127.Savannah\_Sparrow & \xmark & \xmark & \xmark \\\hline
130.Tree\_Sparrow &  \xmark & \cmark & \cmark  \\\hline
132.White\_crowned\_Sparrow & \xmark & \xmark & \xmark \\\hline
136.Barn\_Swallow & \xmark & \xmark & \xmark \\\hline
138.Tree\_Swallow & \cmark & \cmark & \cmark  \\\hline
139.Scarlet\_Tanager & \xmark & \cmark & \cmark \\\hline
143.Caspian\_Tern &  \xmark & \xmark & \cmark  \\\hline
148.Green\_tailed\_Towhee & \xmark & \xmark & \cmark  \\\hline 
156.White\_eyed\_Vireo &  \xmark & \xmark & \cmark  \\\hline 
157.Yellow\_throated\_Vireo & \xmark & \cmark & \cmark \\\hline
161.Blue\_winged\_Warbler &  \xmark & \xmark & \cmark  \\\hline 
163.Cape\_May\_Warbler & \xmark & \xmark & \xmark \\\hline
164.Cerulean\_Warbler & \cmark & \cmark & \cmark  \\\hline
165.Chestnut\_sided\_Warbler & \xmark & \xmark & \cmark  \\\hline 
168.Kentucky\_Warbler & \xmark & \xmark & \cmark  \\\hline 
169.Magnolia\_Warbler & \xmark & \cmark & \cmark  \\\hline
173.Orange\_crowned\_Warbler & \cmark & \cmark & \cmark \\\hline
180.Wilson\_Warbler & \cmark & \cmark & \cmark  \\\hline
188.Pileated\_Woodpecker & \xmark & \xmark & \cmark  \\\hline 
190.Red\_cockaded\_Woodpecker & \cmark & \cmark & \cmark  \\\hline
191.Red\_headed\_Woodpecker & \xmark & \cmark & \cmark  \\\hline
200.Common\_Yellowthroat & \cmark & \cmark & \cmark  \\\hline

\end{supertabular}
\begin{supertabular}{|p{.26\textwidth} H | p{.05\textwidth}  H |} 
\hline
\multicolumn{4}{|c|}{\bf OZSL : SUN \cite{SUN}} \\
\textit{Class Name} & \textit{20-80} & \textit{unseen} & \textit{80-20} \\ \hline
alley & \xmark & \xmark & \cmark  \\\hline
archive & \xmark & \xmark & \xmark \\\hline
arena\_basketball & \cmark & \cmark & \cmark  \\\hline
artists\_loft & \xmark & \cmark & \cmark  \\\hline
auditorium & \xmark & \cmark & \cmark  \\\hline
ballroom & \cmark & \cmark & \cmark  \\\hline
bank\_vault & \xmark & \cmark & \cmark  \\\hline
batting\_cage\_outdoor & \xmark & \cmark & \cmark  \\\hline
bazaar\_indoor & \xmark & \xmark & \cmark  \\\hline
bazaar\_outdoor & \xmark & \xmark & \cmark  \\\hline
betting\_shop & \cmark & \cmark & \cmark  \\\hline
bog & \xmark & \xmark & \cmark  \\\hline
bow\_window\_indoor & \xmark & \cmark & \cmark  \\\hline
bow\_window\_outdoor & \xmark & \cmark & \cmark  \\\hline
brewery\_indoor & \xmark & \xmark & \xmark \\\hline
brewery\_outdoor & \xmark & \xmark & \cmark  \\\hline
bus\_depot\_outdoor & \xmark & \xmark & \cmark  \\\hline
car\_interior\_frontseat & \cmark & \cmark & \cmark  \\\hline
casino\_outdoor & \xmark & \xmark & \cmark  \\\hline
chemistry\_lab & \xmark & \cmark & \cmark  \\\hline
church\_indoor & \xmark & \cmark & \cmark  \\\hline
church\_outdoor & \xmark & \cmark & \cmark  \\\hline
doorway\_indoor & \xmark & \xmark & \cmark  \\\hline
elevator\_interior & \xmark & \cmark & \cmark  \\\hline
excavation & \cmark & \cmark & \cmark  \\\hline
exhibition\_hall & \xmark & \xmark & \cmark  \\\hline
field\_cultivated & \xmark & \cmark & \cmark  \\\hline
firing\_range\_indoor &  \xmark & \xmark & \cmark  \\\hline
fishpond & \xmark & \cmark & \cmark  \\\hline
galley & \xmark & \cmark & \cmark  \\\hline
geodesic\_dome\_indoor &  \xmark & \xmark & \cmark  \\\hline
hangar\_indoor & \xmark & \xmark & \cmark  \\\hline
hoodoo & \cmark & \cmark & \cmark  \\\hline
hotel\_room & \xmark & \xmark & \cmark  \\\hline
ice\_shelf & \xmark & \xmark & \cmark  \\\hline
jacuzzi\_indoor & \xmark & \xmark & \cmark  \\\hline
japanese\_garden & \xmark & \cmark & \cmark  \\\hline
lawn & \xmark & \cmark & \cmark  \\\hline
monastery\_outdoor & \cmark & \cmark & \cmark  \\\hline
mosque\_indoor & \cmark & \cmark & \cmark  \\\hline
motel & \xmark & \xmark & \cmark  \\\hline
observatory\_outdoor &  \cmark & \cmark & \cmark  \\\hline
parking\_lot & \xmark & \cmark & \cmark  \\\hline
piano\_store & \xmark & \xmark & \xmark \\\hline
promenade\_deck & \xmark & \cmark & \cmark  \\\hline
pub\_indoor & \cmark & \cmark & \cmark  \\\hline
racecourse & \xmark & \cmark & \cmark  \\\hline
rectory & \xmark & \xmark & \cmark  \\\hline
sandbox & \cmark & \cmark & \cmark  \\\hline
savanna & \cmark & \cmark & \cmark  \\\hline
ski\_resort & \xmark & \cmark & \cmark  \\\hline
temple\_south\_asia & \xmark &  \xmark & \cmark  \\\hline
theater\_indoor\_seats &  \cmark & \cmark & \cmark  \\\hline
ticket\_booth & \xmark & \cmark & \cmark  \\\hline
trading\_floor & \xmark & \cmark & \cmark  \\\hline
train\_station\_platform &  \xmark & \cmark & \cmark  \\\hline
tundra & \xmark & \xmark & \cmark  \\\hline
tunnel\_road\_outdoor & \xmark & \cmark & \cmark  \\\hline
volleyball\_court\_outdoor &  \xmark & \cmark & \cmark  \\\hline
workshop & \cmark & \cmark & \cmark  \\\hline
wrestling\_ring\_indoor & \xmark & \xmark & \cmark  \\\hline
yard & \xmark & \xmark & \cmark  \\\hline
ziggurat & \xmark & \xmark & \cmark  \\\hline
\end{supertabular}

\begin{supertabular}{|p{.26\textwidth} H | p{.05\textwidth}  H |} 
\hline
\multicolumn{4}{|c|}{\bf OZSL : FLO \cite{FLO}} \\
\textit{Class Name} & \textit{20-80} & \textit{unseen} & \textit{80-20} \\ \hline
Bird\_of\_paradise & \xmark & \xmark & \cmark  \\\hline
Balloon\_flower & \xmark & \xmark & \cmark  \\\hline
Artichoke & \xmark & \cmark & \cmark  \\\hline
Alpine\_sea\_holly &  \cmark & \cmark & \cmark  \\\hline
Barbeton\_daisy & \xmark & \xmark & \xmark  \\\hline
Bolero\_deep\_blue &  \cmark & \cmark & \cmark  \\\hline
Buttercup & \xmark & \cmark & \cmark  \\\hline
Bishop\_of\_llandaff &  \cmark & \cmark & \cmark  \\\hline
Black\_eyed\_susan & \xmark & \cmark & \cmark  \\\hline
Californian\_poppy &  \cmark & \cmark & \cmark  \\\hline
Bearded\_iris & \xmark & \xmark & \cmark  \\\hline
Azalea & \xmark & \cmark & \cmark  \\\hline
Anthurium & \xmark & \xmark & \cmark  \\\hline
Bee\_balm & \xmark & \xmark & \cmark  \\\hline
Ball\_moss & \xmark & \xmark & \xmark  \\\hline
Bougainvillea & \xmark & \xmark & \xmark  \\\hline
Camelia & \xmark & \cmark & \cmark  \\\hline
Bromelia & \xmark & \cmark & \cmark  \\\hline
Blanket\_flower & \xmark & \xmark & \cmark  \\\hline
Blackberry\_lily & \xmark & \xmark & \xmark  \\\hline
\end{supertabular}
\end{document}